\documentclass[10pt,journal,compsoc]{IEEEtran}

\usepackage{times}

\usepackage{soul}
\usepackage{url}
\usepackage[hidelinks]{hyperref}
\usepackage{bm}
\usepackage[utf8]{inputenc}
\usepackage[small]{caption}
\usepackage{graphicx}
\usepackage{amsfonts}
\usepackage{booktabs}
\usepackage{color}
\usepackage{forloop,tabu}
\usepackage{tabu}
\usepackage{amsmath}
\usepackage{amsfonts} 
\usepackage{algorithmic}
\usepackage{algorithm}
\usepackage{array}
\usepackage{stfloats}
\usepackage{url}
\usepackage{booktabs}
\usepackage{multirow}
\usepackage{bm}
\usepackage{color}
\usepackage{ragged2e}

\begin{document}
\title{Weakly-supervised Domain Adaption for Aspect Extraction via Multi-level Interaction Transfer}

\author{Tao~Liang, Wenya~Wang, and Fengmao~Lv
  \thanks{This work was conducted when T. Liang worked as a research assistant at Southwestern University of Finance and Economics, co-supervised by F. Lv and W. Wang. All the authors contribute equally to this work. (Corresponding author: Fengmao Lv.)}
    \thanks{T. Liang is with Center of Statistical Research \& School of Statistics, Southwestern University of Finance and Economics, China, and Content Platform of Center, Tencent, China (e-mail: taoliangdpg@126.com).}
    \thanks{W. Wang is with  School of Computer Science and Engineering, Nanyang Technological University, Singapore (e-mail: wangwy@ntu.edu.sg).}
    \thanks{F. Lv is with Center of Statistical Research \& School of Statistics, Southwestern University of Finance and Economics, China (e-mail: fengmaolv@126.com).}
}

\IEEEtitleabstractindextext{
  \begin{abstract}
  \justifying
Fine-grained aspect extraction is an essential sub-task in aspect based opinion analysis. It aims to identify the aspect terms (a.k.a. opinion targets) of a product or service in each sentence. However, expensive annotation process is usually involved to acquire sufficient token-level labels for each domain. To address this limitation, some previous works propose domain adaptation strategies to transfer knowledge from a sufficiently labeled source domain to unlabeled target domains. But due to both the difficulty of fine-grained prediction problems and the large domain gap between domains, the performance remains unsatisfactory. This work conducts a pioneer study on leveraging sentence-level aspect category labels that can be usually available in commercial services like review sites to promote token-level transfer for the extraction purpose. Specifically, the aspect category information is used to construct pivot knowledge for transfer with  assumption that the interactions between  sentence-level aspect category and  token-level aspect terms are invariant across domains. To this end, we propose a novel multi-level reconstruction mechanism that aligns both the fine-grained and coarse-grained information in multiple levels of abstractions. Comprehensive experiments demonstrate that our approach can fully utilize sentence-level aspect category labels to improve cross-domain aspect extraction with a large performance gain.
\end{abstract}
  }

\maketitle

\IEEEdisplaynontitleabstractindextext

\IEEEpeerreviewmaketitle

\IEEEraisesectionheading{\section{Introduction}\label{sec:introduction}}
\IEEEPARstart{D}{ifferent} from sentence- or document-level sentiment analysis, fine-grained opinion mining aims to produce structured opinion summarizations as a final goal given an opinionated corpus~\cite{Pang:2008:OMS:1454711.1454712}, which is crucial for information extraction. It involves several subtasks including aspect term extraction~\cite{Hu04,Qiu11}, aspect categorization~\cite{Lu09,Hima14}, aspect-based sentiment prediction~\cite{Jiang11,dong14,Ma17} and opinion summarization~\cite{Lu09}. Among these subtasks, aspect term extraction is a key problem in fine-grained opinion mining, which is an essential step towards structured opinion summary generation for the review corpus. It involves the identification of explicit aspect terms in each sentence from online user reviews. Specifically, an aspect term usually refers to an attribute of a product or service. For instance, in the sentence: ``I like the service staff'',  \emph{service staff} is the aspect term that needs to be extracted. Existing works for this task include unsupervised methods that either rely on pattern mining and manually-designed rules~\cite{Hu04,Qiu11} or topic modeling~\cite{Titov08,Lu09}. However, rule-based methods are not flexible to deal with informal texts of user reviews. Topic modeling only works for grouping of aspect terms instead of explicit extraction. Supervised methods were also investigated that treat the extraction task as a sequence labeling problem using either graphical models~\cite{jakob2010extracting,jin2009novel} or deep learning~\cite{yin2016unsupervised,li2017deep,wang2017coupled}. However, the performance of these models depends on the availability of sufficient token-level labels which are usually extremely expensive to obtain.

To alleviate this problem,  a few domain adaptation strategies have been proposed to transfer knowledge from a sufficiently labeled source domain (e.g., \emph{laptop}) to an unlabeled target domain (e.g., \emph{restaurant}) \cite{wang2019transferable,ding2017recurrent,wang2018recursive,li2012cross,jakob2010extracting}. In general, the source and the target domains are bridged together through exploring the prior knowledge of syntactic structure of each sentence \cite{ding2017recurrent,li2012cross} or transferring the implicit memory interactions of aspect and opinion terms within each sentence \cite{wang2019transferable}. Nevertheless, due to both the difficulty of fine-grained prediction problems and the large domain shift between the source and the target domains, the performance of the existing domain adaptation approaches is far from satisfactory. Even if both aspect term and opinion term labels are provided, which is extremely labor-intensive, the additional performance gain brought by the opinion term labels is trivial~\cite{wang2019transferable}. 

Compared to the costly fine-grained token-level annotations, i.e., aspect term labels, the coarse-grained annotations, e.g., sentence-level aspect categories, are more easily accessible. The aspect category usually refers to a pre-defined aspect topic that indicates an overall aspect of a product. In most cases, such aspect category information is readily available in the commercial services like review sites or social media which usually defines a fixed set of aspect topics towards the products or events in a particular domain (such as \textit{food}, \textit{service}, \textit{price} etc. in the restaurant domain and \textit{display}, \textit{keyboard}, \textit{mouse}, etc. in the laptop domain). The aspect category labels for each sentence is provided by the users who tag their comments based on these aspect categories. As a result, a mass collection of reviews tagged with aspect category labels are easily attained in e-commerce, e.g., Amazon and Booking.com~\cite{li18}.  In contrast, token-level annotations are rarely seen and the existing annotated dataset is usually small.

Intuitively, the sentence-level aspect category annotations can be used to improve the performance of aspect term extraction in a weakly supervised manner. This can be reasonable since we observe  a strong association between aspect categories and specific aspect terms: the sentence-level category information provides important clue to identify the exact aspect terms. On the other hand, the specific aspect terms in each sentence are strong indicators to inform the sentence-level aspect categories. For example, the aspect term \textit{service staff} most probably infers the service category. Furthermore, the intensive interaction between coarse-grained categories and fine-grained aspect terms is invariant across domains, which could be utilized to transfer knowledge from the source to the target domain. 

Motivated by the above observations, in this work, we propose a novel transfer setting for aspect extraction based on \emph{weakly-supervised domain adaptation}, where sentence-level aspect category annotations are available for both source and target domains, whereas the aspect term labels are only available in the source domain. To effectively leverage these sentence-level labels for token-level transfer, we propose a \textbf{M}ulti-level \textbf{S}entence-\textbf{W}ord \textbf{I}nteraction \textbf{T}ransfer (MSWIT) model. Specifically, 
MSWIT constructs 2 interactive components: 1) the \textbf{fine-grained} component generates fine-grained high-level feature representation for each word encoding aspect-relevant information for aspect terms extraction. 2) The \textbf{coarse-grained} component produces a coarse-grained sentence vector representation incorporating the overall aspect category information through attending to relevant aspect words within each sentence. Though with different granularity, both of these 2 components exploit important features corresponding to aspect information within a sentence, which could be aligned through a reconstruction mechanism. We adopt a multi-level reconstruction mechanism to reconstruct the fine-grained features towards the coarse-grained representations in varying abstraction levels. Knowledge transfer occurs by sharing the reconstruction module across different domains, with the intuition that the interactions between fine-grained and coarse-grained features are invariant regardless of different domains. To this end, the reconstruction loss in the target domain promotes effective information flow from the given category labels to the absent aspect term labels. 
Although topic models associate informative words with sentence-level topics via a generative process, the association is usually reflected through co-occurrence statistics without considering high-level features~\cite{DBLP:journals/tkde/SoleimaniM15}. Comprehensive experiments over three benchmark datasets clearly demonstrate that our approach can fully utilize the sentence-level aspect category labels to significantly improve the performance of domain adaptation approaches for aspect term extraction. 

Overall, the contributions of this work are mainly three-fold:
\begin{itemize}
\item We conduct a pioneer exploration on weakly supervised domain adaptation for aspect term extraction. Specifically, we effectively leverage the sentence-level aspect category labels that are much more easily accessible than the token-level labels to promote the transferability of aspect information across two different domains.
\vspace{0.1cm}
\item We propose a novel model called MSWIT for weakly-supervised domain adaptation. MSWIT mainly focuses on transferring the implicit interactions between sentence categories and aspect terms through a multi-level reconstruction mechanism.  
\vspace{0.1cm}
\item We conduct comprehensive experiments on six transfer settings over three different domains to verify our proposal. The experimental results clearly demonstrate the effectiveness of our approach.
\end{itemize}

The rest of the paper is organized as follows. Section 2 reviews the existing works on both domain adaptation and aspect term extraction. Section 3 introduces the motivation of this work. Section 4 presents our approach to weakly-supervised domain adaptation for aspect extraction. Section 5 displays our experimental results. Lastly, Section 6 summarizes our proposal with a conclusion.

\section{Related Work}

\subsection{Aspect  Extraction}

Aspect extraction has been extensively studied in the past few years. Early works adopted an unsupervised strategy e.g., association rule mining~\cite{Hu04}, constructing rules that specifies syntactic relations among aspect words and opinion words~\cite{Qiu11,Zhuang06,Pop05}. However, these approaches require external linguistic resources to construct rules and are inflexible for informal texts. Topic modeling has also been proposed for unsupervised extraction~\cite{Titov08,Lu09,Mei07,he17,Zhang10,fangtao10,chen14}, but restrain themselves with misalignment between topics and their semantic meanings. Moreover, topic modeling mostly works for coarse-grained categories, but perform worse on aspect extraction for each sentence. 

On the other hand, supervised approaches have been mostly presented recently by treating the extraction task as a sequence labeling problem and learns a model either through human-engineered features with probabilistic graphical models~\cite{Li10,jin2009novel,jakob2010extracting,Ma10} to exploit contextual and syntactically-related interactions, or deep learning that generates high-level features considering sequential correlations~\cite{liu2015fine,hu18} via recurrent neural networks or convolutional neural networks, dependency-tree-based correlations~\cite{yin2016unsupervised,Wang16} via recursive neural networks and non-local correlations~\cite{li2017deep,wang2017coupled} via attention models. Another line of works applied a word alignment model to capture
opinion relations among a sentence~\cite{Liu12,Liu13}. However, these supervised learning models highly depend on the availability of token-level label annotations which is costly to obtain.

 \subsection{Domain Adaptation} 
 
Domain adaptation is an essential problem in transfer learning.  In general, machine learning works under the assumption that the training data and the testing data come from an identical distribution. However, this assumption is usually not satisfied in transfer learning where distribution shift usually exists between the source and  target domains~\cite{DBLP:journals/tkde/PanY10}.  Domain adaptation strategies aim to bridge the distribution shift and generalize the model's prediction ability from the source domain to the target domain. The existing domain adaptation methods mainly focus on image recognition or document  sentiment classification~\cite{DBLP:conf/icml/LongC0J15,DBLP:journals/corr/abs-1912-11976,DBLP:journals/jmlr/GaninUAGLLML16,DBLP:conf/iclr/ZellingerGLNS17,DBLP:conf/icml/LongZ0J17}. Their key insight is to learn transferable high-level features through aligning the distributions of the source and target domains. Typical approaches towards distribution alignment include domain adversarial training~\cite{DBLP:journals/jmlr/GaninUAGLLML16,DBLP:conf/icml/LongZ0J17} and moment matching~\cite{DBLP:conf/icml/LongC0J15,DBLP:journals/corr/abs-1912-11976,DBLP:conf/iclr/ZellingerGLNS17}.

 \subsection{Domain Adaptation for Aspect Extraction} 
 
 To avoid the labor-intensive labeling for aspect term extraction, a few works propose to use unsupervised domain adaptation methods to transfer knowledge from a labeled source domain to an unsupervised target domain. To this end, Jakob et al. firstly utilized the transferable CRF model to incorporate  non-lexical features, which are usually shared by different domains~\cite{jakob2010extracting}. Li et al. constructed transferable relations among the aspect and opinion words with a boosting strategy to expand the aspect terms in the target domain~\cite{li2012cross}. 
 
 Recently, deep learning approaches have also been proposed to learn domain-invariant representations for each word. Ding et al. and Wang et al. modeled the transferable relation between aspect and opinion words through dependency relations~\cite{ding2017recurrent,wang2018recursive,WangP19}. Recently, Wang et al. proposed an interactive memory network that is transferable without any domain knowledge~\cite{wang2019transferable}. However, the performance of the existing domain adaptation methods is still far from satisfactory due to both the difficulty of fine-grained prediction problems and the clear domain mismatch between the source and  target domains.  Different from previous approaches, this work proposes to utilize sentence-level aspect category labels that are usually readily available in e-commerce to promote token-level transfer for effective aspect term extraction.

Prior to this work, Li et al. proposed to utilize coarse-grained labels to facilitate fine-grained prediction~\cite{li18}. However, their work only dealt with aspect-oriented sentiment prediction and requires a few fine-grained labels in the target domain. Different from~\cite{li18} which proposed to bridge the gap between different domains via distribution alignment, our approach reduces the domain discrepancy by constructing transferrable sentence-word interaction with the intuition that the interactive relation between fine-grained and coarse-grained features is invariant regardless of different domains.

\section{Motivation}

\subsection{Domain-invariant Interaction} 

With only sufficient fine-grained annotations in the source domain, the huge domain gap between different domains makes it infeasible to directly apply a learned model from the source domain to the target domain. Although a few transfer learning methods have been proposed, most of them rely on domain-invariant syntactic structures which are sometimes inaccurate and infeasible for informal texts like customer reviews. Hence, it is still difficult to achieve acceptable performances in the target domain. To tackle this issue, we propose a weakly-supervised approach by providing sentence-level aspect category labels in both the source and target domains, which can be usually more easily accessible compared with the token-level aspect term labels.

It should be noted that the sentence-level aspect categories are strongly correlated with the aspect terms within each sentence. On the one hand, the sentence-level  category information can provide important clues to identify the exact aspect terms. For example, in the sentence ``\textit{I hate the staff}'', the aspect term is \textit{staff} and the sentence-level aspect category is \textit{service}. Given \textit{service} as the aspect category, it becomes easier to identify \textit{staff} as the specific aspect term by comparing each word's relevance score towards the overall category. On the other hand, the specific aspect terms in each sentence are strong indicators to the sentence-level aspect categories. Moreover, this kind of correlation is invariant across different domains, which can be utilized as a bridge to transfer knowledge from the source domain to facilitate  token-level extraction in the target domain.

\subsection{Multi-level Interaction Transfer} 

To this end, a crucial point lies in how to model the invariant word-sentence correlations. To achieve that, we generate two sentence vectors from different granularities. The first vector learns a coarse-grained aspect category representation for the whole sentence by learning to attend to most informative words. The second vector is aggregated from fine-grained word features through aspect-related word transformations in a multi-level hierarchy. Though these two feature vectors are produced through different granularities, they focus on similar information, i.e., the aspects mentioned in the text. With this intuition, a decoder network with a multi-level reconstruction mechanism is adopted to align the coarse-grained and fine-grained feature representations. 

The transferability is reflected by sharing this reconstruction module across the source and  target domains. Hence, the correlation between aspect categories and specific aspect terms learned from the source domain can be transferred to the target domain, through which we can infer the explicit aspect terms for each sentence given the category labels.  

\section{Approach}

\subsection{Problem Statement}

Formally, the task of fine-grained aspect extraction aims to identify the aspect terms within each sentence consisting of a sequence of tokens $\textbf{x}=\{x_1,x_2,...,x_n\}$. To model it as a sequence labeling problem, a sequence of corresponding output labels is produced, denoted by $\textbf{y}=\{y_1,y_2,...,y_n\}$, where $y_i \in \{{\rm BA},{\rm IA} , {\rm N}\}$. Specifically, ${\rm BA}$, ${\rm IA}$, and ${\rm N}$ represent ``beginning of an aspect term'', ``inside an aspect term'' and ``not an aspect term'', respectively. In the cross-domain setting, a source domain $\mathcal{D}^S=\{(\textbf{x}_i^{S},\textbf{y}_i^{S})\}_{i=1}^{N_S}$ annotated with token-level labels is provided in order to transfer knowledge to a new target domain $\mathcal{D}^T=\{\textbf{x}_i^{T}\}_{i=1}^{N_T}$ without any labels on aspect terms. 

Under the setting of weakly-supervised domain adaptation, the sentence-level aspect category labels are provided in both domains, denoted by $\textbf{z}_i^{S} \in \{0,1\}^{|\mathcal{C}^S|}$ for the source domain and $\textbf{z}_i^{T} \in \{0,1\}^{|\mathcal{C}^T|}$ for the target domain, where $\mathcal{C}^m=\{C^m_1,C^m_2,...,C^m_{|\mathcal{C}^m|}\}$ indicates the set of pre-defined aspect categories in domain $m\in \{S, T\}$. By incorporating the sentence-level aspect category labels, the given corpus involves $\mathcal{D}^S=\{(\textbf{x}_i^{S},\textbf{y}_i^{S},\textbf{z}_i^{S})\}_{i=1}^{N_S}$ for the source domain and  $\mathcal{D}^T=\{(\textbf{x}_i^{T},\textbf{z}_i^{T})\}_{i=1}^{N_T}$ for the target domain. The final goal is to produce $\textbf{y}^{T}$ as the sequence of aspect term labels in the target domain.

\begin{figure*}[t]
\centering
\includegraphics[width=0.90\textwidth,trim=0 0 20 0,clip]{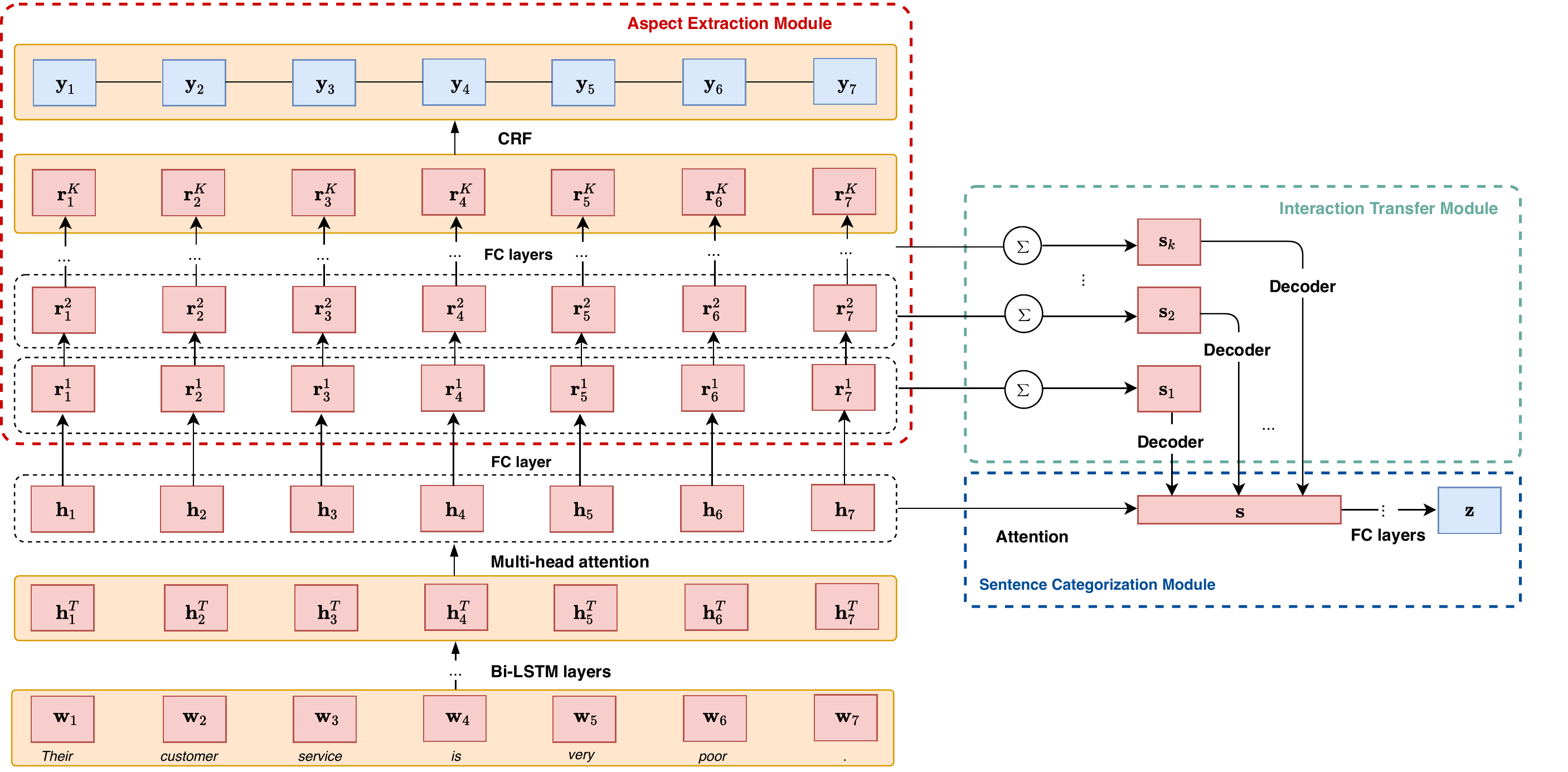}\\
\vspace{0.18cm}
\caption{ The overall architecture of our model. The aspect extraction module produces the token-level aspect term labels. The sentence categorization module produces the sentence-level aspect category labels. The interaction transfer module contains multiple reconstruction units to associate the token-level and sentence-level information. The sentence categorization module and the aspect extraction module share a common multi-layer Bi-LSTM for transforming input word vectors $\mathbf{w}$ to high-level context-sensitive features $\mathbf{h}^T$ and a multi-head attention layer for encoding complex word dependencies to produce hidden vector $\mathbf{h}$.}
\label{model}
\end{figure*}

\subsection{Model Overview}

The architecture of MSWIT is shown in Figure \ref{model}. Overall, MSWIT consists of three components: 1) a token-level \textbf{Aspect Extraction Module} for the source domain; 2) a multi-label \textbf{Sentence Categorization Module} for both source and target domains; 3) an \textbf{Interaction Transfer Module} for learning domain-invariant correlations between fine-grained and coarse-grained feature representations. The aspect extraction module and sentence categorization module share a common multi-layer Bi-LSTM for transforming the input word embeddings $\mathbf{w}$ to high-level context-sensitive features $\mathbf{h}^T$, and a multi-head attention layer for encoding complex word dependencies to produce the final hidden representations $\mathbf{h}$. Then the token-level aspect extraction module further transforms $\mathbf{h}$ via multiple fully-connected layers and a Conditional Random Field (CRF)~\cite{crf01icml} layer on top to produce the final structured predictions for a sequence. On the other hand, the sentence-level categorization module produces a single vector $\mathbf{s}$ to represent the general aspect category-related information which is further fed into $|\mathcal{C}^m|$ binary classifiers to predict the existence for each aspect category. To capture the interactions between the token-level and sentence-level modules, the interaction transfer module is implemented to associate the high-level word vectors for aspect term extraction with the sentence vectors for aspect categorization through multiple reconstruction layers.

 \subsection{Aspect Extraction \& Categorization} 
 
 To exploit rich feature representation for each word, we incorporate both sequential and non-local aspect-related word dependencies through a multi-layer Bi-LSTM followed by a multi-head self-attention model. Specifically, given an input sequence $\{x_1,...,x_n\}$ with word embeddings $\{\textbf{w}_1,...,\textbf{w}_n\}$, the Bi-LSTM model exploits sequential context interactions among a sentence to produce a hidden representation for each word $\mathbf{h}^t_i=[\overrightarrow{\mathbf{h}}^t_i;\overleftarrow{\mathbf{h}}^t_i]$, where
\begin{eqnarray}
\overrightarrow{\mathbf{h}}^t_i  &=&  f(\mathbf{h}^{t-1}_i, \overrightarrow{\mathbf{h}}^t_{i-1};\theta), \nonumber \ \ \ \ \ \  \ \ \   \ \ \  \ \ \  \\
 \hspace{2mm} \overleftarrow{\mathbf{h}}^t_i  &=&  f(\mathbf{h}^{t-1}_i, \overleftarrow{\mathbf{h}}^t_{i+1};\theta). \ \ \  \ \ \  \ \ \  \ \ \  \ \ \ 
\end{eqnarray}
Here $t$ denotes the $t$th Bi-LSTM layer with $\mathbf{h}^0_i$ being the input word vector $\mathbf{w}_i$. Denote the hidden representation obtained at the last Bi-LSTM layer as $\mathbf{h}^T_i$ for each word, then the multi-head self-attention model produces an attention weight vector $\bm{\alpha}_i$ and a transformed hidden vector $\mathbf{h}_i$ for each word with the following procedure:
\begin{eqnarray}
\ \ \ \mathbf{h}_i &=& [\bar{\mathbf{h}}^1_i;\bar{\mathbf{h}}^2_i;,,,;\bar{\mathbf{h}}^M_i] \\
\ \ \ \bar{\mathbf{h}}^m_i &=& \sum_{j=1}^n \bm{\alpha}^m_{ij} (\mathbf{W}^m_V \mathbf{h}^T_j), \\
 \ \ \   \bm{\alpha}^m_{i} &=& \textup{softmax} (\frac{(\mathbf{W}^m_Q \mathbf{h}^T_i)(\mathbf{W}^m_K \mathbf{H}^T)}{\sqrt{d}}). 
\end{eqnarray}
Here $\bar{\mathbf{h}}^m_i$ indicates the transformation from an attention head $m\in\{1,...,M\}$ and $\bm{\alpha}^m_{i}$ indicates the attention weight vector measuring the interaction between $x_i$ and all the other tokens for attention head $m$. $\mathbf{H}^T$ is a matrix consisting of the output $\mathbf{h}^T_i$ from the multi-layer Bi-LSTM as column vector. $d$ is the hidden-layer dimension.

The output $\mathbf{h}_i$ from the attention layer can be fed into 2 modules. Specifically, for the aspect extraction module, we apply multiple fully-connected layers to transform $\mathbf{h}_i$ to $\mathbf{r}^k_i$ in the $k$th FC (fully-connected) layer. We denote $\mathbf{r}^k_i$ as
\begin{equation}
\mathbf{r}^k_i=\textup{FC}^{e}(\mathbf{r}^{k-1}_i) \hspace{2mm}\mbox{with}\hspace{2mm} \mathbf{r}^0_i=\mathbf{h}_i.
\end{equation}
The final token-level labels are produced by feeding $\mathbf{r}^{K}_i$ for each word to the CRF model, where $K$ is the number of FC layers preceding the CRF layer. The CRF model is able to exploit sequential label dependencies that are beneficial for aspect extraction. The final token-level label sequence is denoted as
\begin{eqnarray}
\hat{\mathbf{y}} = \textup{argmax}_{\mathbf{y}} \frac{\exp(\phi(\mathbf{r}^{K},\mathbf{y}))}{\sum_{\mathbf{y}'}\exp(\phi(\mathbf{r}^{K},\mathbf{y}'))},
\end{eqnarray}
where $\phi$ indicates the score function in a CRF model for the given input $(\mathbf{r}^K, \mathbf{y})$ and $\mathbf{y}'$ ranges over all possible label sequences.

On the other hand, for the sentence categorization module, the multi-head attention output is passed to a general attention layer to generate the attention weight  $\bm{\alpha}^g_i$ and the sentence vector $\mathbf{s}$ by attending to the most informative words towards the aspect category, represented as follows: 
\begin{eqnarray}
\textbf{s} = \sum_{i=1}^{n} \bm{\alpha}_i^{g} \cdot \textbf{h}_i \hspace{2mm}\mbox{with}\hspace{2mm}
\bm{\alpha}_i^g  = {\rm softmax}(\textbf{W} \cdot \textbf{h}_i).
\end{eqnarray}
We use $\mathbf{s}$ as the input for $|\mathcal{C}^m|$ ($m\in\{S, T\}$ indicates the domain) fully-connected neural networks for binary predictions corresponding to $|\mathcal{C}^m|$ different aspect categories: $\hat{\mathbf{z}}_{[j]}=\sigma(\textup{FC}^{c}_j(\mathbf{s}))$, where $j\in\{1,...,|\mathcal{C}^m|\}$. $\sigma$ is the sigmoid function that generates probabilistic outputs.

 \subsection{Interaction Transfer Module} 
The key insight of our approach is to learn the invariant correlations between the coarse-grained categories and the fine-grained aspect terms.  It is clear that the separate modules for aspect extraction and category prediction fail to exploit the correlations between these two tasks. To associate these two modules, we propose a multi-level reconstruction mechanism to exchange information across granularities and learn the domain-invariant word-sentence correlations. 

\begin{table*}[h!]
\fontsize{9.0}{14.5} \selectfont
\centering
\caption{Data description for different domains.}
\label{dataset}
\renewcommand\arraystretch{1.0}
\begin{tabular}{m{40pt}<{\centering}| m{55pt}<{\centering}| m{240pt}|m{35pt}<{\centering}|  m{30pt}<{\centering} |m{30pt}<{\centering} }
\toprule
\hline
 Dataset & Description & \centering Category & Sentences & Training & Testing\\
\hline
LAP & Laptop &   \emph{laptop}, \emph{os}, \emph{display}, \emph{multimedia\_devices},  \emph{mouse},  \emph{keyboard}, \emph{battery}, \emph{graphics},  \emph{hard\_disc},  \emph{software},  \emph{support}, \emph{company} & 3,845 & 2,861 & 984 \\
\hline
RES & Restaurant & \emph{service}, \emph{anecdotes}, \emph{ambience}, \emph{price},   \ \emph{food} & 5,841 & 4,381 & 1,460 \\
\hline
DEV & Digital Device & \emph{device}, \emph{display}, \emph{phone}, \emph{radio}, \emph{software}, \emph{hardware}, \emph{button}, \emph{battery} & 3,836 & 2,838 & 998\\
\hline
\end{tabular}
\end{table*}

To be specific, the token-level aspect term extraction module produces fine-grained word vectors encoding aspect-relevant information. On the other hand, the sentence-level aspect categorization module produces coarse-grained sentence vectors by attending to the informative aspect words. With the same focus on the aspect information in each sentence, the high-level word features and sentence vector should be aligned. This is achieved through a decoder which tries to decode the word features with close proximity to the sentence vector $\mathbf{s}$. Furthermore, multi-level reconstruction is implemented to ensure the alignment with different levels of abstractions. Formally speaking, we first aggregate the high-level feature representation $\mathbf{r}^k_i$ at $k$th FC layer for each word to form a fine-grained sentence vector: 
\begin{eqnarray}
\mathbf{s}_k = \sum_{i=1}^{n}\mathbf{r}^k_i. 
\end{eqnarray}
Denote these sentence vectors for all the $K$ FC layers as $\{\mathbf{s}_1,...,\mathbf{s}_{K}\}$. We can build correlations between coarse-grained and fine-grained information through reconstruction with multiple levels in the hierarchy.
The reconstruction module is implemented through the L2 norm:
\begin{eqnarray}
\ \ \ \ \ \ d(\mathbf{s}, \mathbf{s}_k) =  \left \| \ \mathbf{s} - \textbf{W}_{k}^{2} \cdot {\rm Relu}(\textbf{W}_{k}^{1} \cdot \textbf{s}_k)\right \|^2.
\end{eqnarray}
Specifically, the decoder contains two fully-connected layers and there are $K$ different decoders indicating different abstraction levels. Under the reconstruction mechanism,  the decoders will focus on learning the underlying interaction between word-level and sentence-level representations. They are shared across different domains with the intuition that the interactions between the fine-grained and coarse-grained information are invariant across different domains. As the source domain has both fine-grained and coarse-grained labels, the sentence-word correlations can be accurately captured within the source domain. With the decoders shared across two domains, the correlations learned from the source domain can be transferred to the target domain. Then the specific aspect terms can be informed in the target domain given the sentence-level aspect category labels.

 \subsection{Training}
 
 From our definition of weakly-supervised domain adaptation, we assume the availability of both aspect-term labels and sentence-level aspect category labels in the source domain $\mathcal{D}^S=\{(\textbf{x}_i^{S},\textbf{y}_i^{S},\textbf{z}_i^{S})\}_{i=1}^{N_S}$, but only sentence-level aspect category labels in the target domain $\mathcal{D}^T=\{(\textbf{x}_i^{T},\textbf{z}_i^{T})\}_{i=1}^{N_T}$. Hence, we can train the  extraction model via a token-level cross-entropy loss in the source domain:
\begin{equation}
    \mathcal{L}_{e} = -\frac{1}{N_S}\sum_{i=1}^{N_S} \sum_{n=1}^{n_i} \mathbf{y}_{i,n}\log\hat{\mathbf{y}}_{i,n},
\end{equation}
where $n_i$ is the token length of $\textbf{y}_i$. The sentence categorization model is trained on both the source and target domains with the multi-label loss:
\begin{eqnarray}
    \mathcal{L}_{c} &=& -\frac{1}{N_S}\sum_{i=1}^{N_S}\sum_{j=1}^{|\mathcal{C}^S|}\mathbf{z}^S_{i[j]}\log\hat{\mathbf{z}}^S_{i[j]} \nonumber \\ 
    && - \frac{1}{N_T}\sum_{i=1}^{N_T}\sum_{j=1}^{|\mathcal{C}^T|}\mathbf{z}^T_{i[j]}\log\hat{\mathbf{z}}^T_{i[j]}.
\end{eqnarray}
The reconstruction loss is computed for both source and target domains as:
\begin{eqnarray}
    \mathcal{L}_{r} &=& \frac{1}{N_S}\sum_{i=1}^{N_S}\sum_{k=1}^{K}d(\mathbf{s}_{i}^{S}, \mathbf{s}^{S}_{i,k}) 
 \nonumber \\ 
 &&    +  \frac{1}{N_T}\sum_{i=1}^{N_T}\sum_{k=1}^{K}d(\mathbf{s}_{i}^{T}, \mathbf{s}^{T}_{i,k}).\nonumber
\end{eqnarray}
The reconstruction loss in the source domain constructs the interactive relations between the fine-grained and coarse-grained information. On the other hand, the reconstruction loss in the target domain promotes effective information flow from the given category labels to the absent aspect term labels.

In conclusion, with the above sub-objectives, our final
objective function is formulated as follows:
\begin{eqnarray}
\mathcal{L} = \mathcal{L}_{e} + \lambda \mathcal{L}_{c} +\beta \mathcal{L}_{r},
\end{eqnarray}
where $\lambda$ and $\beta$ are hyper-parameters that weigh the importance of the corresponding modules.

\section{Experiments}
\subsection{Datasets} 
Our experiments are conducted over three customer review datasets, including {laptop}, {restaurant} and {digital device}. Specifically, the {laptop} domain is composed of the laptop reviews from SemEval 2014 task 4 subtask 1 \cite{pontiki2014semeval}. The {restaurant} domain is a combination of the restaurant reviews from SemEval 2014 task 4 subtask 1 and  SemEval 2015 task 12 subtask 1 \cite{pontiki2015semeval}. The reviews in the {digital device} domain are from \cite{Hu04}. We refer the readers to \cite{wang2018recursive} for more details about the datasets. 
The restaurant and laptop domains already contain sentence-level category labels. Moreover, we conduct an extra step to annotate the aspect  category labels for each sentence from the digital device domain\footnote{ We first select 1000 random sentences from DEV and use them to pre-define a set of aspect categories for DEV. Then we annotate the sentences from the pre-defined set. If a new aspect category appears in the annotation process, we'll add it into the pre-defined set. Two annotators were involved in the annotation. Cross-check is conducted to correct wrong annotations. The inter-annotator agreement  is 86.2\%. The annotated datasets will be released in the final version.}. Table \ref{dataset} displays the data description for each domain. Following \cite{wang2018recursive}, we randomly split the datasets (with 3/4 for training and 1/4 for testing) for five times and report the average results over those data splits.

\subsection{Implementation details} 

We implement our method using the PyTorch framework. The word embeddings of the input sentences are obtained through Skip-Gram pre-trained on the English wiki corpus~\cite{mikolov2013efficient}. The dimension of the word embeddings is 300. In order to show the robustness of our approach, we use the identical hyper-parameters in all the settings.  Three sentence reconstruction modules are used to construct the domain-invariant sentence-word  relations. Moreover, 3 Bi-LSTM layers and 4 fully-connected layers are used in the actual model implementation. The size of the Bi-LSTM and FC layers are set to 400 and 512, respectively. In our experiments, the trade-off parameter of the sentence categorization loss ($\lambda$) is set to 0.4 and that of the multi-level reconstruction loss ($\beta$) is set to 0.8. To train our model, we adopt Adam with  learning rate of 0.008 as the optimizer. The batch size is set to 64 and the  epoch number is set to 20. Additionally, we adopt the dropout strategy to alleviate the overfitting problem, with the dropout rate set to 0.45. To evaluate the extraction performance, we adopt F1 score as the metric. The exact multi-word term match is counted as a true positive. All the experiments are conducted on a GeForce GTX 1080 Ti GPU. 
 
 \begin{table*}[t!]
\fontsize{9.0}{13} \selectfont
\centering
\caption{Comparisons with different domain adaptation baselines in terms of the F1 score.  $\rm{RNSCN}^{\dagger}$, $\rm{TIMN}^{\dagger}$ and MSWIT are trained under the weakly-supervised domain adaptation setting. The other baselines are trained under the unsupervised domain adaptation setting.}
\label{compare}
\renewcommand\arraystretch{1.0}
\begin{tabular}{m{57pt}<{\centering} |m{52pt}<{\centering}|m{55pt}<{\centering}|m{52pt}<{\centering}|m{54pt}<{\centering}|m{55pt}<{\centering}|m{54pt}<{\centering}}
\toprule
\hline
Models & LAP $\rightarrow$ RES & LAP $\rightarrow$ DEV & RES $\rightarrow$ LAP & RES $\rightarrow$ DEV & DEV $\rightarrow$ LAP & DEV $\rightarrow$ RES\\
\hline
Source-Only &50.05&39.47&38.17&32.41&43.91&48.27\\
\hline
TCRF & 28.19 & 29.96 & 19.72 & 21.07 & 24.22 & 6.59\\
\hline
RNNAL & 48.10&31.25&33.66&33.20&34.74&47.97\\
\hline
RAP & 46.90&34.54&25.92&22.63&28.22&45.44 \\
\hline
RNSCN&52.91&40.42&40.43&35.10&51.41&48.36\\
\hline
TIMN&54.12&38.63&43.68&35.45&52.46&53.82\\
\hline
$\rm{RNSCN}^{\dagger}$&52.97&39.37&41.34&35.86&53.54&50.23\\
\hline
$\rm{TIMN}^{\dagger}$&55.13&40.72&44.71&35.37&54.42&51.57\\
\hline
MSWIT&\textbf{61.73}&\textbf{46.17}&\textbf{46.81}&\textbf{41.19}&\textbf{56.77}&\textbf{56.19}\\
\hline
\end{tabular}
\end{table*}

\begin{table*}[t!]
\fontsize{9.0}{13} \selectfont
\centering
\caption{Ablation study for the contribution of each module in terms of the F1 score. The notations -SCM and -ITM indicate removing  the Sentence  Categorization Module (SCM) and the Interaction Transfer Module (ITM), respectively. The notation -ITMs indicates training  ITM  without using the source data.}
\label{ablation}
\renewcommand\arraystretch{1.0}
\begin{tabular}{m{57pt}<{\centering} |m{52pt}<{\centering}|m{55pt}<{\centering}|m{52pt}<{\centering}|m{54pt}<{\centering}|m{55pt}<{\centering}|m{54pt}<{\centering}}
\toprule
\hline
Models & LAP $\rightarrow$ RES & LAP $\rightarrow$ DEV & RES $\rightarrow$ LAP & RES $\rightarrow$ DEV & DEV $\rightarrow$ LAP & DEV $\rightarrow$ RES\\
\hline
None & 50.05&39.47&38.17&32.41&43.91&48.27\\
\hline
 -SCM & 53.45&41.97&41.67&34.59&45.73&50.26\\
\hline
 -ITM & 54.38&43.18&42.13&37.07&51.13&52.57\\
\hline
 -ITMs &59.71&44.17&44.49&37.69&53.29&52.64\\
\hline
MSWIT &\textbf{61.73}&\textbf{46.17}&\textbf{46.81}&\textbf{41.19}&\textbf{56.77}&\textbf{56.19}\\
\hline
\end{tabular}
\end{table*}
 
\subsection{Baselines} 
To demonstrate the effectiveness of our approach, we conduct comprehensive comparisons with different models, including both  unsupervised domain adaptation baselines and  weakly-supervised domain adaptation baselines, which are clearly listed as follows:
\begin{itemize}
\item $\textbf{Source-Only}$: The vanilla aspect term extraction model trained with only the source domain data. To some extent, the source-only result reflects the lower bound of the extraction performance that a model can ever achieve in a domain adaptation setting.
\vspace{0.1cm}
\item \textbf{TCRF}: Transferable Conditional Random Field proposed in \cite{jakob2010extracting}. It treats the non-lexical features (e.g., POS tags and dependency relations) shared by different domains as the pivot information to promote the transferability of CRF. 
\vspace{0.1cm}
\item \textbf{RNNAL}: Recurrent Neural Network with Auxiliary Labels from \cite{ding2017recurrent}. It leverages  domain-independent rules to generate auxiliary labels for transfer.
\vspace{0.1cm}
\item \textbf{RAP}: Relational Adaptive bootstraPping proposed in \cite{li2012cross}. It iteratively generates high-confidence aspect seeds at the target domain and expands them by exploiting general syntactic relation patterns.
\vspace{0.1cm}
\item \textbf{RNSCN}:  Recursive Neural Structural Correspondence Network proposed in \cite{wang2018recursive}. It treats the syntactic relations between adjacent words as invariant pivot information across domains, in order to construct the structural correspondences and reduce the token-level domain shift.
\vspace{0.1cm}
\item \textbf{TIMN}: Transferable Interactive Memory Network from \cite{wang2019transferable}. It exploits local and global memory interactions to explore both the intra-correlations and inter-connections within sentences for aligning source and target spaces.
\item \textbf{MSWIT}:  Multi-level Sentence-Word Interaction  proposed in this work.
\end{itemize}

\noindent For fair comparison, we further incorporate the sentence-level aspect category labels into the state-of-the-art models (e.g., RNSCN and TIMN) through equipping them by the sentence categorization module with the multi-label loss $\mathcal{L}_C$, which is the naive way to leverage weak supervision. The corresponding models are denoted as $\rm{RNSCN}^{\dagger}$ and $\rm{TIMN}^{\dagger}$, respectively. Note that for RAP, RNSCN and TIMN, the opinion terms of the source domain need to be additionally annotated, which will significantly increase the labeling consumption. In contrast,  the sentence-level aspect category labels used in MSWIT are usually available in  commercial services such as review sites or social media.

 \begin{table*}[t!]
\fontsize{9.0}{13} \selectfont
\centering
\caption{Comparisons with different numbers of  reconstruction layers in terms of the F1 score. When the number of reconstruction modules is set to 0, the fine-grained and coarse-grained modules are trained separately.}
\label{num}
\renewcommand\arraystretch{1.0}
\begin{tabular}{m{57pt}<{\centering} |m{52pt}<{\centering}|m{55pt}<{\centering}|m{52pt}<{\centering}|m{54pt}<{\centering}|m{55pt}<{\centering}|m{54pt}<{\centering}}
\toprule
\hline
Layers & LAP $\rightarrow$ RES & LAP $\rightarrow$ DEV & RES $\rightarrow$ LAP & RES $\rightarrow$ DEV & DEV $\rightarrow$ LAP & DEV $\rightarrow$ RES\\
\hline
0 & 54.38&43.18&42.13&37.07&51.13&52.57\\
\hline
1 & 56.18&44.52&45.19&38.27&53.56&53.17\\
\hline
2 & 59.04&46.01&46.23&39.86&55.19&55.76\\
\hline
3&\textbf{61.73}&\textbf{46.17}&\textbf{46.81}&41.19&\textbf{56.77}&\textbf{56.19}\\
\hline
4&59.37&45.93&42.73&\textbf{42.71}&55.47&53.18
\\
\hline
\end{tabular}
\end{table*}

\begin{table*}[t!]
\fontsize{9.0}{13} \selectfont
\centering
\caption{Comparisons with different variations of the backbone network in terms of the F1-score.  When the number of attention head is set to 0,  the model does not include the self-attention layer. }
\label{num_lstm}
\renewcommand\arraystretch{0.95}
\begin{tabu}{m{52pt}<{\centering} |m{18pt}<{\centering}|m{50pt}<{\centering}|m{50pt}<{\centering}|m{50pt}<{\centering}|m{50pt}<{\centering}|m{50pt}<{\centering}|m{50pt}<{\centering}}
\toprule
\hline
\multicolumn{2}{c|}{Variations} & LAP $\rightarrow$ RES & LAP $\rightarrow$ DEV & RES $\rightarrow$ LAP & RES $\rightarrow$ DEV & DEV $\rightarrow$ LAP & DEV $\rightarrow$ RES\\
\hline
\multirow{5}{*}{{LSTM  layer} } & 0 & 58.71&43.03&43.87&37.73&53.17&54.39
\\
\cline{2-8}
&1 & 60.48&44.83&45.19&39.36&54.91&55.78
\\
\cline{2-8}
&2 & 60.31&45.59&46.77&40.54&55.71&\textbf{56.37}
\\
\cline{2-8}
&3& \textbf{61.73}&\textbf{46.17}&\textbf{46.81}&41.19&\textbf{56.77}&56.19
\\
\cline{2-8}
&4&60.19&44.96&45.27&\textbf{42.13}&55.17&55.43
\\
\cline{1-8}
\multirow{4}{*}{{Attention  head} } & 0 & 60.14 & 45.21 & 46.83 & 40.01 & 53.72 & 56.41
\\
\cline{2-8}
&1 & 61.17&45.69&{47.12}&40.72&54.93&\textbf{56.49}
\\
\cline{2-8}
&4 & 61.39&46.03&\textbf{47.29}&40.91&56.18&55.79
\\
\cline{2-8}
&8 & \textbf{61.73}&\textbf{46.17}&{46.81}&\textbf{41.19}&\textbf{56.77}&{56.19}
\\
\hline
\end{tabu}
\end{table*}

\subsection{Performance Comparison} 

Table \ref{compare} displays the comparisons of different baselines. The results are averaged over 5 different random splits of  data. TCRF and RAP are based on traditional feature-engineering strategies. Compared with deep  models, they cannot learn high-level representations that contain transferable factors, and hence obtain relatively poor cross-domain performance. Both RNNAL and RNSCN heavily rely on the  accuracy of pre-mined knowledge. The inaccurate syntactic relations pre-generated by the external linguistic tools will affect the extraction performance, or even cause the problem of negative transfer, which is clearly revealed in the comparison between
 Source-Only and RNNAL. Although the most recent method TIMN gets rid of the dependency over external resources through exploiting local and global memory interactions to capture the domain-independent syntactic relations within sentences, it needs to additionally annotate the opinion terms token by token, which significantly increases the labeling consumption. In contrast, our approach turns to use the sentence-level aspect category labels instead, which can be  usually more easily accessible than the token-level labels. It is clear that our proposed approach can obtain significantly better domain adaptation performance compared with TIMN. 

Compared with $\rm{RNSCN}^{\dagger}$ and $\rm{TIMN}^{\dagger}$, the larger performance improvement obtained by our proposed MSWIT clearly demonstrates the effectiveness of our approach for leveraging the sentence-level aspect category labels to promote the transferability of token-level labels between the source and target domains. Moreover, we also conduct t-test with the null hypothesis being ``the baseline model and the proposed model have the same domain adaptation performance''. The null hypothesis is rejected when $p<0.01$. Notably, the performance improvement over $\rm{RNSCN}^{\dagger}$ is  significant for all the settings. Compared with $\rm{TIMN}^{\dagger}$, the extraction performance for the LAP $\rightarrow$  RES, RES $\rightarrow$  DEV and DEV $\rightarrow$  RES settings are significant. 

\begin{figure*}[h!]
\centering
{\includegraphics[width=0.98\textwidth, height=0.21\textwidth, trim=80 11 115 40,clip]{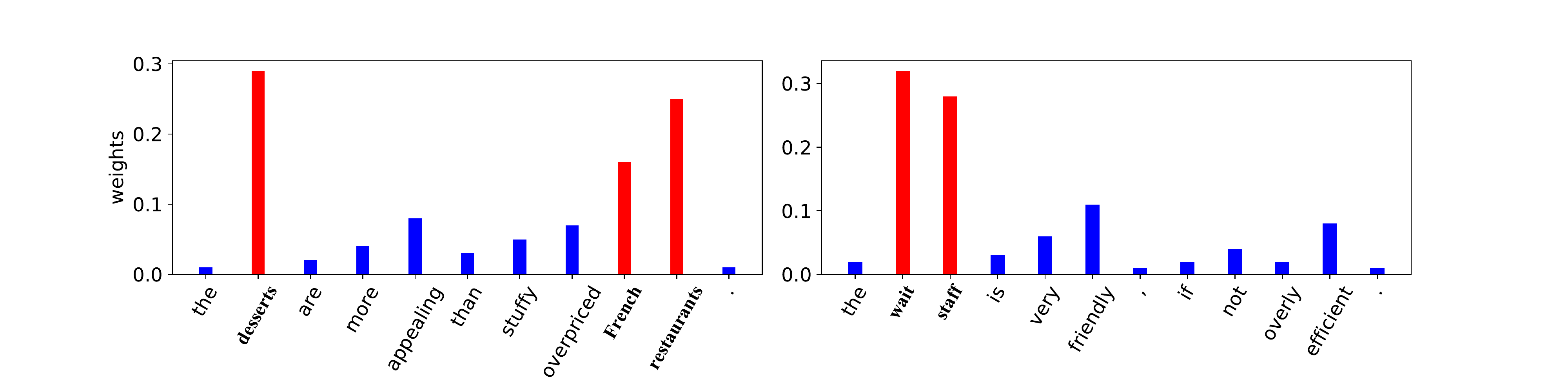}}
\caption{Visualization of attention weights $\alpha_{i}^{g}$ for different tokens within a sequence. The true aspect tokens are displayed in bold text.}
\label{vis}
\end{figure*}

\subsection{Ablation Study}

As displayed in Table \ref{ablation}, we further conduct the ablation study to demonstrate the contribution of each module in MSWIT. The first row represents the vanilla aspect term extraction model trained with only the source domain data. The notation ``-SCM'' in the second row indicates removing the \textit{Sentence Categorization Module} from MSWIT. As a result, the sentence vector $\mathbf{s}$ does not have any supervision. We can see that although weak supervision is not utilized, the reconstruction module can still correlate sentences with words to some extent. Therefore, our proposed MSWIT can also be suitable for the unsupervised domain adaptation setting. However, as no sentence categorization loss enforces the sentence vectors to learn the semantics of aspect categories, the interactive relation obtained by the reconstruction module alone is relatively poor and hence leads to only small improvements. The notations ``-ITM'' and ``-ITMs'' denote the model without \textit{Interactive Transfer Module} and without source data in the interactive transfer module, respectively. When the interactive transfer module is removed, the sentence categorization branch induces attentions that are beneficial to locating the corresponding aspect terms, and hence makes a contribution to aspect extraction. With the addition of the \textit{Interactive Transfer Module} only applied on the target domain, the performance is further improved. This shows the ability of the reconstruction model for associating word-level and category-level information. However, the performance is still far from satisfactory when compared with the proposed model, in which a better interactive transfer module is learned utilising all the category labels from both the source and target domains. The clear performance difference between ``-ITMs'' and ``MSWIT'' indicates that the sentence-word interactive relations are mainly captured by the source domain which is then transferred to the target domain. This observation confirms our hypothesis that the sentence-word interactive relations are invariant across different domains.

\vspace{0.1cm}

\noindent \textbf{Multi-level reconstruction mechanism.}  MSWIT model consists of multiple reconstruction layers to correlate the word vectors for aspect extraction with the sentence vectors for aspect categorization. The multi-level architecture is crucial for capturing the interactive relation between sentence-level aspect categories and aspect terms in different levels of abstraction. To demonstrate this, we conduct comparisons with different number of reconstruction layers as displayed in Table \ref{num}. It is clear that increasing the number of reconstruction layers from 1 to 3 leads to gradual improvements for aspect extraction. However, performance drop appears when 4 reconstruction layers are adopted. Therefore, it is inferred that the interactive relations between sentence categories and aspect terms can be sufficiently captured with 3 reconstruction levels. More reconstruction layers will increase the number of parameters to be learnt and lead to overfitting. 

\vspace{0.1cm}

\noindent \textbf{Variations of the backbone network.} We also conduct comparisons with different variations of the backbone network in Table \ref{num_lstm}. From the results obtained by varying the number of LSTM layers, it is clear that multiple Bi-LSTM layers are necessary for obtaining powerful representations capturing the contextual correlations within a sentence. Similar pattern occurs for the number of attention heads. When the number of attention heads is set to 0, it corresponds to the model where the self-attention model is removed. It is obvious that the self-attention model with multiple attention heads can lead to better performance since it is able to encode complex word dependencies with multiple dimensions. In general, the backbone network with 3 LSTM layers and 8 attention heads provide the best performance. 

\subsection{Analysis}

For qualitative analysis, we demonstrate the effect of knowledge transfer in terms of normalized attention weights in the target domain. As shown in Figure~\ref{vis}, we plot the normalized attention score for each word of 2 sentences in the target domain. Clearly, the domain-invariant reconstruction module promotes the attention model in the target domain to select the most informative words (aspect words) for aspect categorization even without aspect term annotations in the training data. 

Our proposed model may produce inaccurate predictions in some situations, which we illustrate as error analysis:
\begin{enumerate}

\item \emph{The \textbf{food} is uniformly exceptional, with a very capable \textbf{kitchen} which will proudly whip up whatever you feel like eating, whether it's on the \underline{\textcolor{red}{\textbf{menu}}} or not.}

\vspace{0.1cm}

\item \emph{They did not have \textbf{mayonnaise}, forgot our \textbf{toast}, left out \textbf{ingredients} (i.e, \textbf{cheese} in an \textbf{omelet}), below hot temperatures and the \textbf{bacon} was so over cooked it crumbled on the \underline{\textcolor{red}{\textbf{plate}}} when you touched it. }

\end{enumerate}
For these two cases,  \textit{menu}  and \textit{plate} are incorrectly extracted as the aspect terms.  These 2 cases reflect one type of error generated from MSWIT, i.e., false positive predictions. Here \textit{menu} and \textit{plate} are both related to the aspect category \textit{food}, but without any opinions. Hence, they should not be extracted as aspect terms. This might be caused by the model trying to associate each category-related word as an aspect word without checking the opinion information. This limitation could be addressed in our future work.

\vspace{0.1cm}

\begin{figure*}[t]
\centering
\includegraphics[width=1.0\textwidth, height=0.49\textwidth,trim=0 0 0 0,clip]{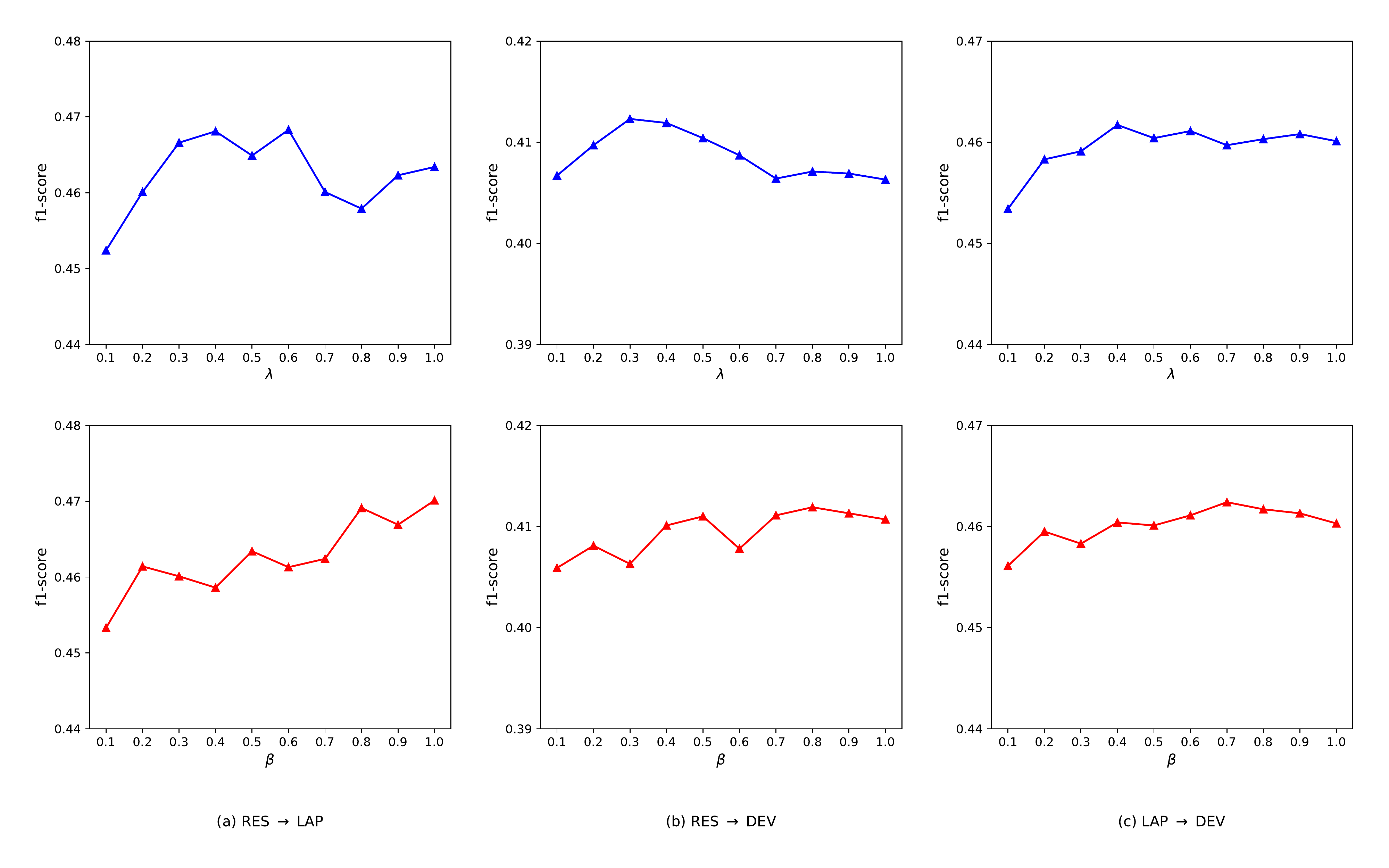}\\
\vspace{-0.4cm}
\caption{Sensitivity analysis for trade-off parameters. The sensitivity analysis is conducted through changing the value of $\lambda$ ($\beta$), while fixing  $\beta$ ($\lambda$) to the value used in the experiments.}
\label{sensitivity}
\end{figure*}

 Finally, to demonstrate the robustness of the proposed model MSWIT, we conduct the sensitivity analysis for hyper-parameters over three  settings as displayed in Figure \ref{sensitivity}. In particular, we conduct experiments varying the trade-off parameters, including $\lambda$ for the multi-label sentence classification loss and $\beta$ for the multi-level reconstruction loss. From Figure \ref{sensitivity}, it is clear that the performance of MSWIT is not sensitive to the values of these trade-off parameters. 

\vspace{0.1cm}

\begin{figure*}[t!]
\centering
\includegraphics[width=1.0\textwidth, height=0.49\textwidth,trim=10 0 10 0,clip]{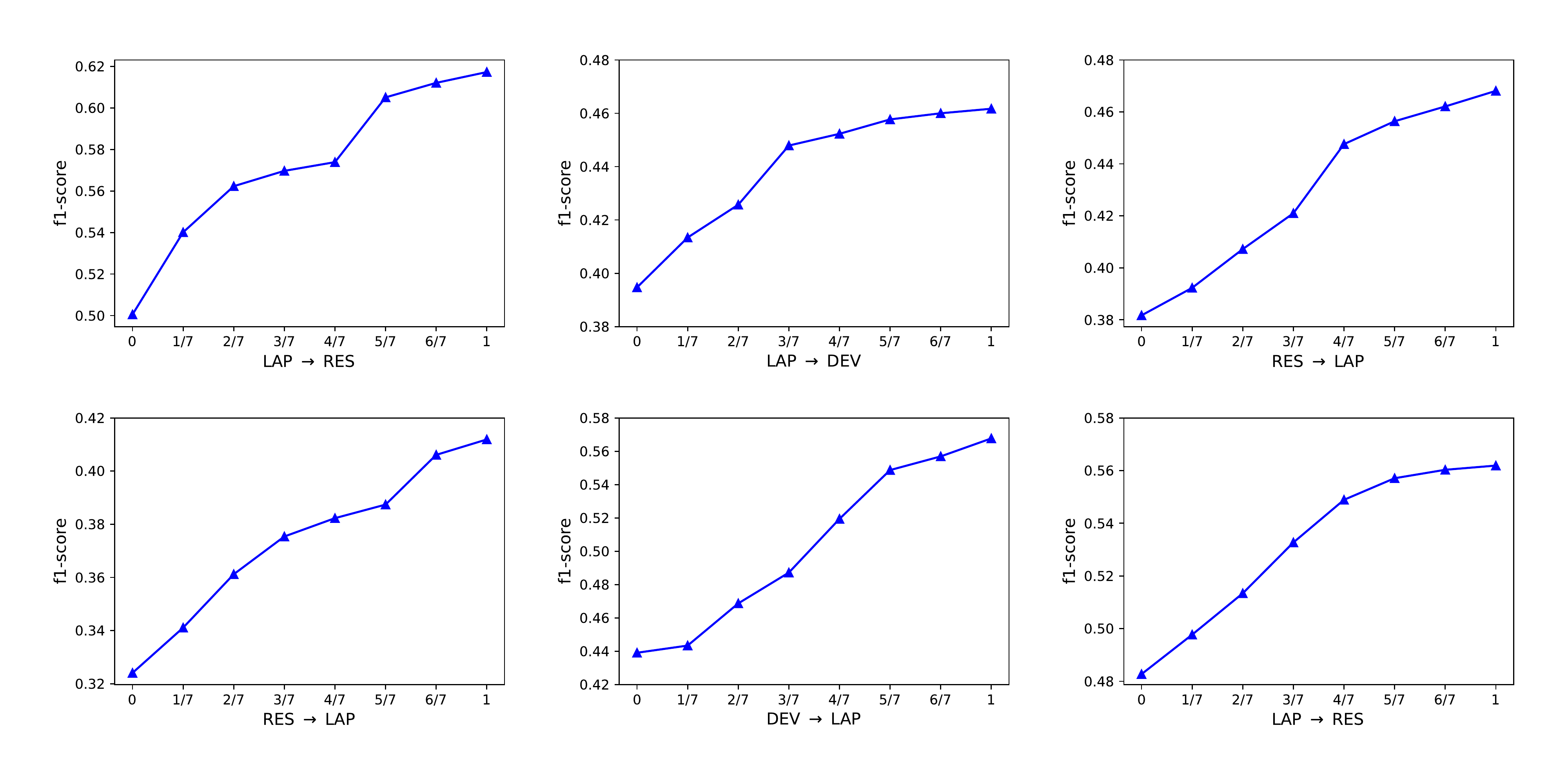}\\
\vspace{-0.5cm}
\caption{F1 score vs proportion of target training data. The horizontal axis indicates the proportion of unlabeled target data.}
\label{transfer}
\end{figure*}

Moreover, we demonstrate our model's transfer ability in Figure~\ref{transfer}. By increasing the number of training data in the target domain, our proposed model shows stable improvements. This demonstrates our model's capability in capturing domain-invariant information to promote extraction in the target domain. 

\section{Conclusion}
In this work, we conduct a pioneer exploration about how to effectively leverage the sentence-level aspect category labels, which are usually much more accessible  than the token-level aspect term labels, to promote the transferability of token-level labels. To tackle this new setting, we propose a novel method named MSWIT, which mainly focuses on transferring the implicit interactions between sentence categories and aspect terms through a multi-level attention-based reconstruction module. We conduct comprehensive experiments to verify our proposal. The experimental results clearly demonstrate that our model can fully utilize weak supervisions to significantly improve the performance of cross-domain aspect extraction.

\ifCLASSOPTIONcompsoc
  \section*{Acknowledgments}
\else
  \section*{Acknowledgment}
\fi
This work is supported by the Lee Kuan Yew Postdoctoral Fellowship of Singapore (No. 200604393R) and the National Natural Science Foundation of China (No. 11829101 and 11931014) and the Fundamental Research Funds for the Central Universities of China  (No. JBK1806002). The authors would like to thank the anonymous reviewers for  their constructive comments.

\ifCLASSOPTIONcaptionsoff
  \newpage
\fi

\bibliographystyle{IEEEtran}
\bibliography{IEEEabrv,mybibfile}

\end{document}